\title{Robustness of sentence length measures in written texts}

\documentclass[3p,preprint,12pt]{elsarticle}
\usepackage{hyperref}
\journal{Physica A}

\begin{document}

\title{Robustness of sentence length measures in written texts}

\author[address]{Denner S. Vieira\corref{mycorrespondingauthor}}
\cortext[mycorrespondingauthor]{Corresponding author}
\ead{dennersvieira@gmail.com}

\author[address]{Sergio Picoli}
\ead{spjunior@dfi.uem.br}
\author[address]{Renio S. Mendes}
\ead{rsmendes@dfi.uem.br}

\address[address]{Departamento de F\'{i}sica, Universidade Estadual de Maring\'a, Avenida Colombo 5790, 87020-900 Maring\'a, Paran\'a, Brazil.}

\date{\today}

\begin{abstract}
Hidden structural patterns in written texts have been subject of considerable research in the last decades. In particular, mapping a text into a time series of sentence lengths is a natural way to investigate text structure. Typically, sentence length have been quantified by using measures based on the number of words and the number of characters, but other variations are possible. To quantify the robustness of different sentence length measures, we analyzed a database containing about five hundred books in English. For each book, we extracted six distinct measures of sentence length, including number of words and number of characters (taking into account lemmatization and stop words removal). We compared these six measures for each book by using \textit{i}) Pearson's coefficient to investigate linear correlations; \textit{ii}) Kolmogorov--Smirnov test to compare distributions; and \textit{iii}) detrended fluctuation analysis (DFA) to quantify auto--correlations. We have found that all six measures exhibit very similar behavior, suggesting that sentence length is a robust measure related to text structure.
\end{abstract}


\maketitle
\section{Introduction}
\label{sec-intro}

Several quantitative patterns have been identified in the structure of written texts. For example, the Zipf's law states that the frequency of a word in a text is inversely proportional to its rank \cite{zipf_psycho-biology_1968,altmann_statistical_2016}. Moreover, it was also observed that punctuation marks also obey this law \cite{kulig_narrative_2017}. Heaps' or Herdan's law states that the vocabulary increases as a power law of a finite sample size of the text \cite{herdan_quantitative_1964,heaps_information_1978}. These laws have been generalized to account for more general patterns \cite{mandelbrot_informational_1953,mandelbrot_simple_1954,mandelbrot_information_1966,petersen_languages_2012,gerlach_stochastic_2013,williams_text_2015,altmann_generalized_2016,ferrer-i-cancho_compression_2016}. Mapping a text into a time series can also reveal hidden structural patterns. 
Methods applied to investigate text structure at word level include investigations on probability distributions \cite{altmann_beyond_2009}, correlations \cite{ebeling_long-range_1995,montemurro_long-range_2002,altmann_origin_2012,ausloos_measuring_2012,ausloos_generalized_2012}, and networks properties \cite{cancho_small_2001,amancio_identification_2012,cong_approaching_2014,kulig_modeling_2015}.
Another way to investigate text structure is by the analysis of sentence lengths. Typically, each sentence carries a full message and transmits an idea in contrast with an isolated word. Therefore, mapping a text into a time series of sentence lengths is a natural way to investigate text structures. 
Recent methods applied to study texts at sentence level include probability distributions \cite{ha_extension_2002,ishida_distributions_2007,yang_long-range_2016,yang_evolution_2017} and correlations \cite{yang_long-range_2016,yang_evolution_2017,altmann_origin_2012,drozdz_quantifying_2016}. In recent studies, sentence length analysis have been related to style and authorship \cite{dzurjuk_sentence_2006,yang_long-range_2016,yang_evolution_2017}. 

In general, sentence lengths have been quantified by the number of words \cite{ha_extension_2002,dzurjuk_sentence_2006,ishida_distributions_2007,drozdz_quantifying_2016} or characters \cite{ausloos_equilibrium_2008,ausloos_punctuation_2010,yang_long-range_2016,yang_evolution_2017}. Also, note that the recurrence time of full stops also quantifies the sentence length \cite{kulig_narrative_2017}. However, there are other possible variations, such as word length or word frequency mappings, where all words are brought to lower case and punctuation marks are removed. Other possible variations are the removal of stop words and the lemmatization \cite{amancio_identification_2012}. In a similar way, the number of characters and variations related to lemmatization and stop words removal could also be considered at sentence level. However, to choose between words or characters to measure a sentence may lead to doubts. For instance, the Japanese language has three alphabets and the number of characters may differ for the same word \cite{ishida_distributions_2007}.

In the present work, we investigated the robustness of distinct measures of sentence lengths. By using a database containing about five hundred books in English, we extracted six distinct measures of sentence length. We employed three widely known tests to compare these six measures for each book: \textit{i}) Pearson's coefficient to investigate linear correlations; \textit{ii}) Kolmogorov--Smirnov test to compare distributions; and \textit{iii}) detrended fluctuation analysis (DFA) to quantify auto--correlations. We have found that all measures considered are very similar, indicating that sentence length is a robust measure related to text structure. This article is structured as follows: in section \ref{sec:data} we describe the data set in which the analysis were executed; in section \ref{sec-results} we performed our analyses comparing the six types of sentence length time series delineated above; and we summarized and discussed our results in section \ref{sec:conc}.

\section{Database}
\label{sec:data}

\begin{figure*}
    \centering
    \includegraphics[width=\linewidth]{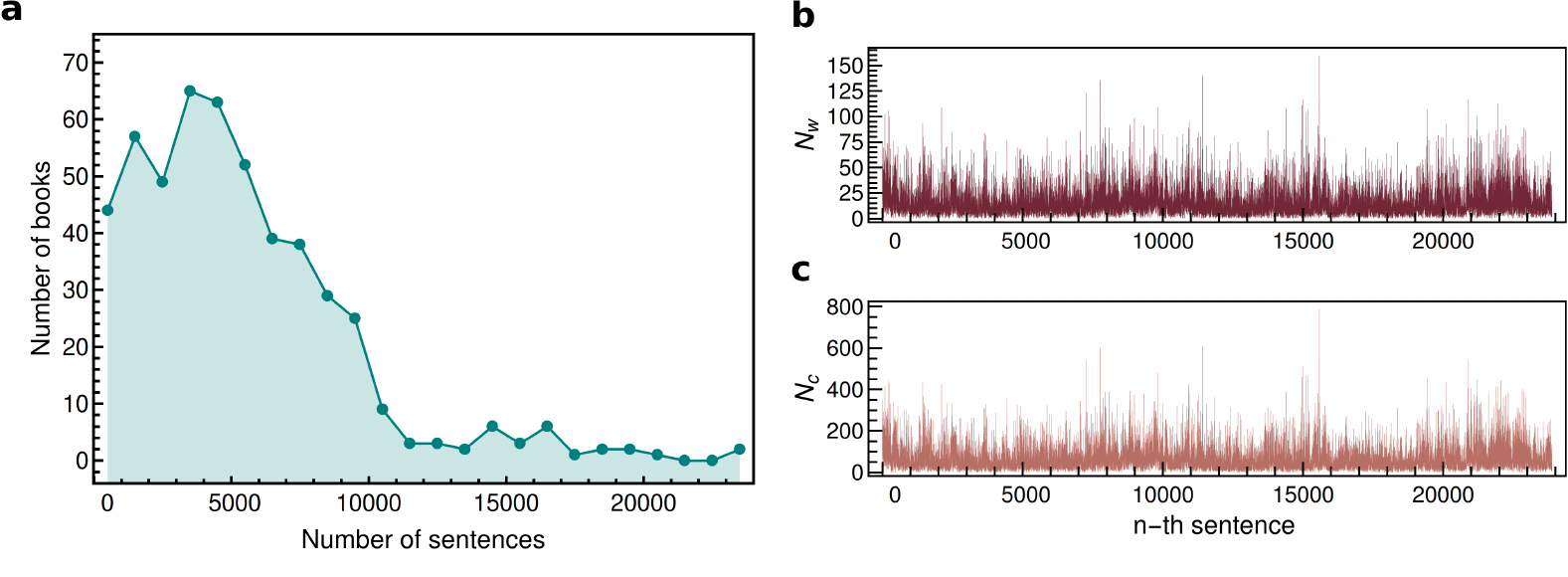}
	\caption{(Color online) \textbf{Data aspects.} Panel \textbf{a} shows the histogram of the number of sentences for all books in our dataset. Panels \textbf{b} and \textbf{c} present the time series of sentence length for the book \textit{The Brothers Karamazov} using words ($N_w$) and characters ($N_c$).}
    \label{fig:sentences_number_per_book}
\end{figure*}

\begin{table*}
\centering
\caption{\textbf{Transformations applied in the texts before measuring sentence length.} To exemplify, we quote a book excerpt from \textit{The Adventures of Sherlock Holmes} written in $1892$ by Arthur Conan Doyle.}
\label{table:text_transformations}
\resizebox{\columnwidth}{!}{%
\begin{tabular}{cccc}
\hline
Original                                                                                                                                                                                                                                                                                       & Lemmatized                                                                                                                                                                                                                                                                             & Non-Stop words                                                                                                                                                                         & Non-Stop words lemmatized                                                                                                                                                          \\ \hline
\begin{tabular}[c]{@{}c@{}}To Sherlock Holmes she\\ is always the woman.\\ I have seldom heard\\ him mention her under\\ any other name. In his eyes\\ she eclipses and predominates\\ the whole of her sex. It was not\\ that he felt any emotion akin\\ to love for Irene Adler.\end{tabular} & \begin{tabular}[c]{@{}c@{}}to sherlock holmes she\\ be always the woman.\\ i have seldom hear\\ him mention her under\\ any other name. in his eye\\ she eclipse and predominate\\ the whole of her sex. it be not\\ that he felt any emotion akin\\ to love for irene adler.\end{tabular} & \begin{tabular}[c]{@{}c@{}}sherlock holmes\\ always woman.\\ seldom heard\\ mention\\ name. eyes\\ eclipses predominates\\ whole sex.\\ felt emotion akin\\ love irene adler.\end{tabular} & \begin{tabular}[c]{@{}c@{}}sherlock holmes\\ always woman.\\ seldom hear\\ mention\\ name. eye\\ eclipse predominate\\ whole sex.\\ felt emotion akin\\ love irene adler.\end{tabular} \\ \hline
\end{tabular}%
}
\end{table*}

Our database consists of all $501$ fiction books written in English, from \textit{The Literature Network} \cite{online_literature}. This database includes famous titles such as \textit{Alice's Adventures in Wonderland} and \textit{Through the Looking Glass} (both written by Lewis Carroll in $1865$ and $1872$, respectively), \textit{Moby Dick} (written in $1851$ by Herman Melville) and \textit{The Adventures of Sherlock Holmes} (written in $1892$ by Arthur Conan Doyle). The time each book was written range from the year $1515$ to $1930$, but most of them ($96.40\%$) were written after $1800$. Here we define the end of a sentence by the occurrence of the following tokens: end stop (``.''), exclamation mark (``!''), and question mark (``?''). With these metric rules, the biggest book in our database is \textit{The Brothers Karamazov} written in $1880$ by Fi\'odor Dostoi\'evski, containing $24,\!186$ sentences while the smallest one is \textit{The Hunting of the Snark} written in $1876$ by Lewis Carroll with 223 sentences. Figure~\ref{fig:sentences_number_per_book}a shows a histogram quantifying the books of our database by the total number of sentences. Most books have around $4,\!500$ sentences.

For each book, we extracted six time series of sentence length measuring the: \textit{i}) number of words, $N_w$; \textit{ii}) number of characters, $N_c$; \textit{iii}) number of characters in lemmatized words, $N_l$; \textit{iv}) number of non--stop words, $N_{Sw}$; \textit{v}) number of characters in non--stop words, $N_{Sc}$; and \textit{vi}) number of characters in non--stop words lemmatized, $N_{Sl}$. Differences between the original text and their variants are exemplified in Table~\ref{table:text_transformations}. Figures~\ref{fig:sentences_number_per_book}b and \ref{fig:sentences_number_per_book}c illustrate two distinct time series of sentence length for a given book, counting words and characters, respectively.

\section{Results}
\label{sec-results}

\subsection{Correlations between mapped time series}
\label{subsec:pearson_results}

\begin{figure*}
    \centering
    \includegraphics[width=\textwidth]{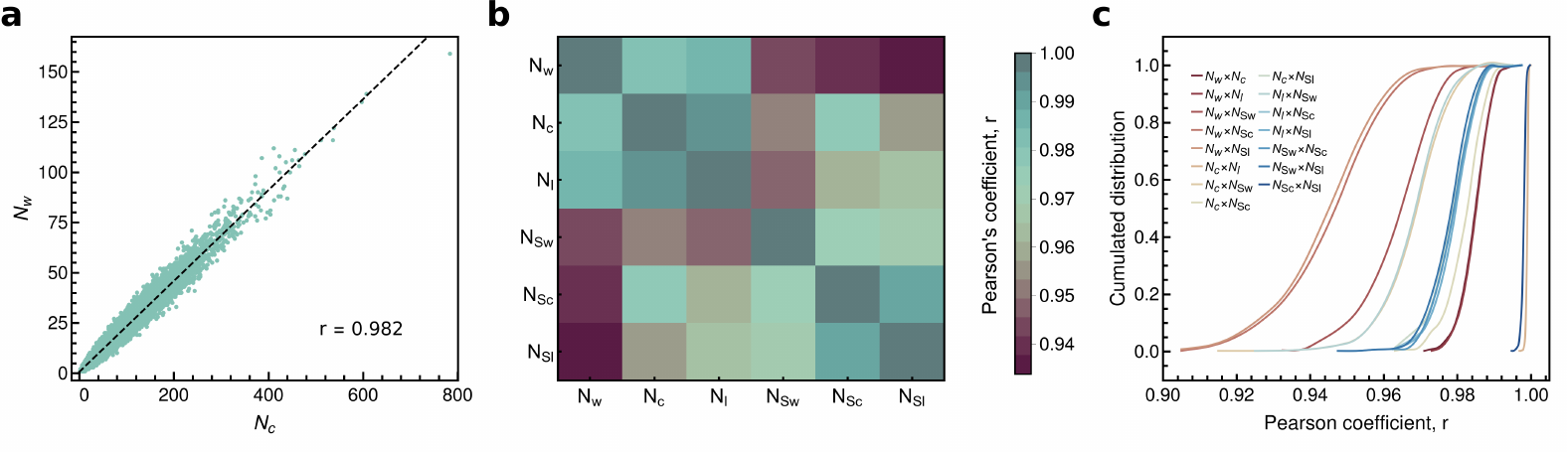}
    \caption{(Color online) \textbf{Linear correlations results.} Panels \textbf{a} and \textbf{b} illustrate the linear correlation between the time series $N_w$ and $N_c$ and the Pearson's coefficient for all comparisons, respectively, for the book \textit{The Brothers Karamazov}. Panel \textbf{c} exhibits the cumulated distributions of Pearson's coefficients of all books for all $7\!,515$ comparisons.}
    \label{fig:pearson_correlation_matrix}
\end{figure*}

To investigate similarities between the six mapped time series, we first analyzed linear correlations via the Pearson's coefficient \cite{sedgwick_pearson_2012}
\begin{equation}
    r = \frac{\displaystyle{\sum\limits^{N}_{i = 1}\left( x_i - \langle X \rangle \right)\left( y_i - \langle Y\rangle \right)}}{\sigma_x \sigma_y},
    \label{eq:Pearson_coef}
\end{equation}
where $\langle X \rangle$ and $\langle Y \rangle$ are the mean values and $\sigma_x$ and $\sigma_y$ are the standard deviations of the sets $X$ and $Y$, respectively, and $N$ is the number of terms of the series. This coefficient lies in the interval $[-1,1]$, where $1$ $(-1)$ means a perfect positive (negative) linear correlation between the two quantities while $0$ means no linear correlation. 

Figure~\ref{fig:pearson_correlation_matrix}a illustrates the correlation between two mapped series, $Y = N_w$ and $X = N_c$, of a book where a monotonic behavior and a strong linear correlation is observed. The Pearson's coefficients for all comparisons for this book is shown in Fig.~\ref{fig:pearson_correlation_matrix}b. According to our analysis, $r > 0.91$. Since we have six time series, there are $6*(6-1)/2 = 15$ comparisons for each book, resulting in $7,\!515$ correlations for all $501$ books. The results above are quite similar to those found for all books for each linear correlation test, as can be seen in the cumulated distributions in Fig.~\ref{fig:pearson_correlation_matrix}c. Among all these coefficients, the weakest and the strongest linear correlations were, respectively $r_{N_w \times N_{Sc}} = 0.91$ (\textit{The Flood}, Emile Zola, $1880$) and $r_{N_{Sc} \times N_{Sl}} = 0.99$ (\textit{A Daughter of Eve}, Honor\'e de Balzac, $1838$). We also notice that the values of $r$ are greater for comparisons among series that count characters. The mean value of all $r$'s was $0.98$. Since $r$ is very close to one, the fluctuation of information due to the transformations has little implications. Due to these results, we estimate that, with good approximation for each book, the terms of one series can be written in terms of the other by a transformation such as
\begin{equation}
    n_c = \alpha_c n_w + \beta_c,
    \label{eq:linear_dependences}
\end{equation}
where $n_c$ ($n_w$) is a term of the time series $N_c$ ($N_w$) and $\alpha_c$ and $\beta_c$ are constants.

Other tests such as the Goodman and Kruskal's $\gamma$ test \cite{goodman_measures_1954}, the Kendall's $\tau$ test \cite{kendall_new_1938}, and the Spearman's test \cite{spearman_proof_1904} were also performed to study the correlations. These tests are based on the null hypothesis that the two distributions are independent. All comparisons for all books rejected the null hypothesis for these tests. 

Before concluding this subsection, we discuss Fig.~\ref{fig:pearson_correlation_matrix}a in connection with the Menzerath--Altmann law. Basically, this law states that the bigger the whole, the smaller its parts and vice-versa. However, an issue about this relationship was addressed in \cite{altmann_prolegomena_1980}: ``Somewhat more problematic is the relation of sentence length to the word length.''  This comment is consistent with the one in \cite{drozdz_quantifying_2016} asserting that the Menzerath--Altmann law does not hold if the sentence length is measured in terms of characters instead of the number of words. In a similar way, the data of Fig.~\ref{fig:pearson_correlation_matrix}a indicates, in the context of this law, that the relationship between the number of words and characters is also problematic.

\subsection{Similarities between distributions}
\label{subsec:correlation_linearity}

\begin{figure*}
    \centering
    \includegraphics[width=\linewidth]{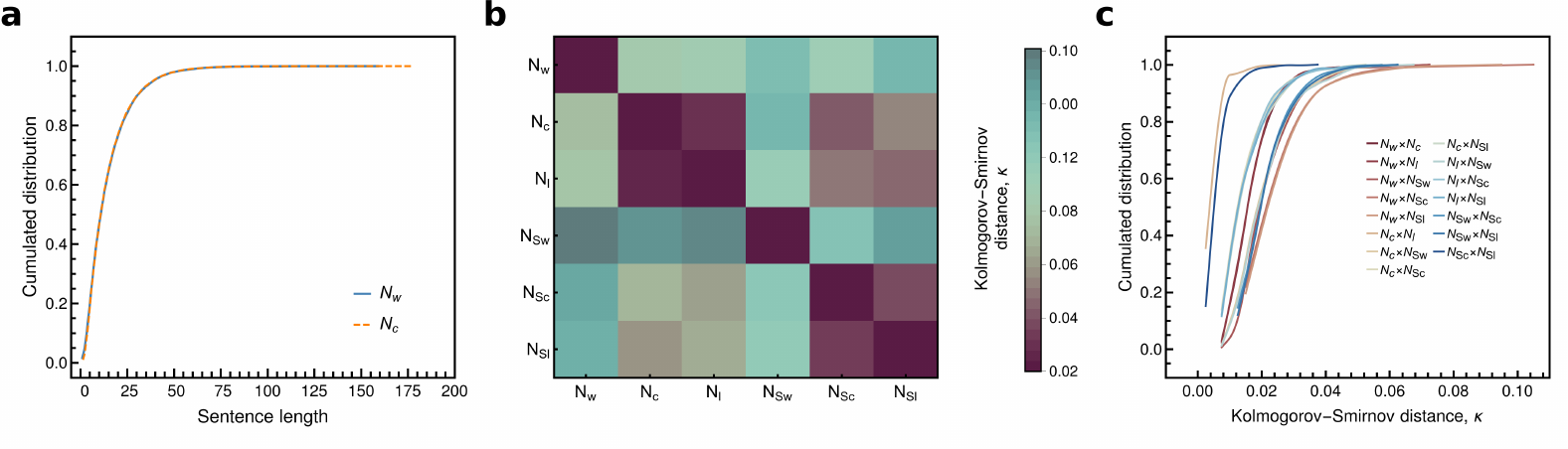}
    \caption{(Color online) \textbf{Distribution comparisons.} Panels \textbf{a} and \textbf{b} illustrate the cumulated distribution for sentence length considering $N_w$ and $N_c$ and the Kolmogorov--Smirnov distances for all comparisons, respectively, for the book \textit{The Brothers Karamazov}. Panel \textbf{c} exhibits the cumulated distributions of all Kolmogorov--Smirnov distances of all books for all $7\!,515$ comparisons.}
    \label{fig:kolmogorov_results}
\end{figure*}

Next, we investigated the distribution of sentence lengths within each book, for each measure of sentence length considered here. This test exploits the distance $\kappa$ between two empirical cumulated distributions $C_1$ and $C_2$:
\begin{equation}
    \kappa = \sup\limits_{x}|C_1(x) - C_2(x)|,
\end{equation}
where $\sup$ is the supremum function; the smaller the $\kappa$ the greater the similarity between the distributions. Typically, the number of characters is greater than the number of words (Figs.~\ref{fig:sentences_number_per_book}b and \ref{fig:sentences_number_per_book}c). Thus, to make a fair comparison, the all time series were normalized by their mean values prior to the KS test. Figure~\ref{fig:kolmogorov_results}a and \ref{fig:kolmogorov_results}b illustrate some results for the book \textit{The Brothers Karamazov}. The former shows the comparison between the cumulated distributions for $N_w$ and $N_c$ after normalization and the latter the Kolmogorov--Smirnov distance ($\kappa$) for all comparisons. Note that the distances found are small ($\kappa \leq 0.12$), implying in the fact that these distributions can be drawn from the same distribution. We verified that $41.77\%$ of these comparisons accepted the null hypothesis that the distributions are drawn from the same distribution, with a \textit{p}--value $\geq 0.01$. When considering the transformation \ref{eq:linear_dependences}, the results for this test (Fig.~\ref{fig:kolmogorov_results}c) were of $89.74\%$ success rate, which implies that the dependence proposed between the series are meaningful. This type of percentage for each type of comparison is exhibited in Table~\ref{table:kolmogorov-pvalues}.

\begin{table}
\centering
\caption{\textbf{Percentage of distributions that can be drawn from the same distributions for each comparison.} The percentage values were obtained using the two--sample Kolmogorov--Smirnov test (\textit{p}--value $\geq 0.01$).}
\label{table:kolmogorov-pvalues}
\resizebox{6cm}{!}{%
\begin{tabular}{lccccc}
\multicolumn{6}{c}{\% \textit{p}-value $\geq 0.01$}                                            \\
\multicolumn{1}{l|}{}         & $N_c$   & $N_l$    & $N_{Sw}$ & $N_{Sc}$ & $N_{Sl}$ \\ \hline
\multicolumn{1}{l|}{$N_w$}    & 95.80   & 96.00    & 80.43    & 71.85    & 67.86    \\
\multicolumn{1}{l|}{$N_c$}    &         & 100.00   & 88.62    & 98.40    & 92.01    \\
\multicolumn{1}{l|}{$N_l$}    &         &          & 92.01    & 99.00    & 98.60    \\
\multicolumn{1}{l|}{$N_{Sw}$} &         &          &          & 90.41    & 88.42    \\
\multicolumn{1}{l|}{$N_{Sc}$} &         &          &          &          & 100.00
\end{tabular}%
}
\end{table}

As an extra remark, we notice that the complementary cumulative distribution (CCDF)  as a function of the number of words, $N_w$, of the data shown in Fig.~\ref{fig:kolmogorov_results}a is consistent with Fig.~6 of \cite{drozdz_quantifying_2016} until $N_w \sim 150$. Explicitly, $CCDF(N_w) = \exp(-\mu N_w^b)$, with $b$ close to unity, decreasing as the  sentence length increases, and $\mu$ is a positive parameter. Because the maximum number of words per sentence of the \textit{The Brothers Karamazov} (the biggest book in our database) goes roughly until $N_w \sim 150$, an additional comparison cannot be performed.

\subsection{Auto--correlations}
\label{subsec:dfa}

\begin{figure*}
    \includegraphics[width=\textwidth]{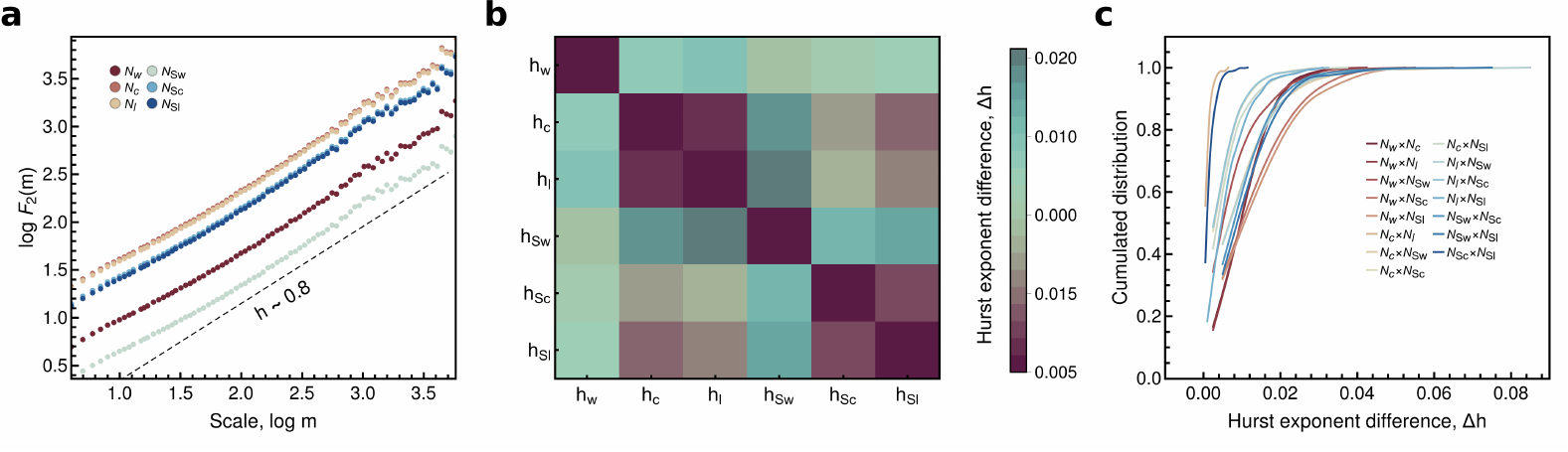}
    \caption{(Color online) \textbf{Auto--correlations.} Panels \textbf{a} and \textbf{b} illustrate the DFA results with $l=1$ for all six series and the absolute difference between the Hurst exponent for all these series, respectively, for the book \textit{The Brothers Karamazov}. Panel \textbf{c} exhibits the cumulated distributions of all Hurst exponents differences of all books for all $7\!,515$ comparisons.}
    \label{fig:hurst}
\end{figure*}

Detrended fluctuation analysis (DFA) have been largely used to study long--range correlations in time series \cite{peng_mosaic_1994,kantelhardt_detecting_2001}. Texts, for instance, were mapped using the sentence length measured as the number of characters and investigated via DFA \cite{grabska-gradzinska_multifractal_2012,yang_long-range_2016,yang_evolution_2017}. To exemplify the method, consider a generic sentence length series $W = \left\lbrace w_1, w_2, \cdots, w_N \right\rbrace $. The first step is to calculate the integrated series $Z = \left\lbrace z_1, z_2, \cdots, z_N \right\rbrace$, where $z_j = \sum_{i = 1}^{j} w_i$, and subdivide it into $s = N/m$ non overlapping partitions with $m$ terms in each window. Each partition is fitted by a polynomial function of degree $l$ and has its fluctuation computed. Averaging the fluctuations over the $s$ segments, we obtain the fluctuation function, $F(m)$, which is usually compared with a power law:
\begin{equation}
    F(m)\!\sim m^h,
\end{equation}
where $h$ is the scale exponent. If the second moment of $Z$ converges, $h$ is the Hurst exponent. For $h>0.5$ ($h<0.5$) the time series has persistent (anti--persistent) long--range correlation and for $h = 0.5$ there is no correlation or it is of short range. It is common to calculate the Hurst $h^*$ for the shuffled version of the time series. It is expected that $h^* = 0.5$.

Using the \textit{The Brothers Karamazov} as example again, Fig.~\ref{fig:hurst}a and \ref{fig:hurst}b illustrate, respectively, $F(m)$ and the differences between the Hurst exponents for all six series. We can infer that $h$ differs very little from one series to another and that their values are close to $0.8$, with $h^*\sim0.5$, implying in long--range correlations. All the series from the other books reflects this behavior ($h\sim0.75$). This result is consistent with  the multifractal analysis performed  in \cite{drozdz_quantifying_2016}. In addition, the values for the difference between Hurst exponents ($\Delta h$) of a given book are shown in Fig.~\ref{fig:hurst}c, where we can observe that the variation of the Hurst exponents is small from one series to another. Also, the standard deviation for $h$ within each book was close to $0.001$. Lastly, no correlations were found between the number of sentences in a text and the Hurst exponent.

\section{Conclusions}
\label{sec:conc}

By considering six ways to map a text into a time series, we have performed tests to gauge the robustness of sentence length measures in written texts. The length units used were characters and words for regular, lemmatized, without stop--words sentences and a combination of the latter two. For each one of the $501$ fiction books in English from our database, strong linear correlations were found when the Pearson's coefficients were calculated obtaining $r\!\sim0.98$ (Fig.~\ref{fig:pearson_correlation_matrix}c). By using this feature, we employed Eq.~\ref{eq:linear_dependences} to map the series into one another in order to compare their distributions using the Kolmogorov--Smirnov test. Among the $7\!,515$ possible comparisons, $89.74\%$ accepted the null hypothesis the they are drawn from the same distribution (Fig.~\ref{fig:kolmogorov_results}c). Due to the interest into long--range correlations in time series mapped from texts, we performed detrended fluctuation analysis to quantify auto--correlations and to study their variations among the constructed series. We have found that the Hurst exponents were very close to each other, $|\Delta h|\!\sim0.001$ (Fig.~\ref{fig:hurst}c), presenting positive long--range correlations ($h\sim0.75$ with $h^*\sim0.5$).

Our results have not pointed into the direction of significant differences among the series considered, indicating a robustness of sentence length measures in written texts in English.

\section*{Acknowledgements}

The authors are grateful for the support of the Brazilian CNPq and CAPES agencies, and the
National Institute of Science and Technology for Complex Systems.

\section*{References}

\bibliographystyle{elsarticle-num}
\bibliography{bibliography.bib}

\end{document}